\documentclass[sigconf,nonacm]{acmart}

\usepackage{tikz}
\usetikzlibrary{arrows.meta,positioning,fit,backgrounds,calc}

\AtBeginDocument{%
  }

\setcopyright{acmlicensed}
\copyrightyear{2018}
\acmYear{2018}
\acmDOI{XXXXXXX.XXXXXXX}
\acmConference[Conference acronym 'XX]{Make sure to enter the correct
  conference title from your rights confirmation email}{June 03--05,
  2018}{Woodstock, NY}
\acmISBN{978-1-4503-XXXX-X/2018/06}

\begin{document}

\title{Where Do LLMs Retrieve Network Information? Identifying the Responsible Heads and Neurons}

\author{Md Abdul Kadir}
\correspondingauthor
\email{abdul.kadir@dfki.de}
\affiliation{%
  \institution{German Research Center for Artificial Intelligence (DFKI)}
  \city{Saarbrücken}
  \country{Germany}}
\affiliation{%
  \institution{University of Oldenburg}
  \city{Oldenburg}
  \country{Germany}}

\author{Md Mohasin Hossain}
\affiliation{%
  \institution{German Research Center for Artificial Intelligence (DFKI)}
  \city{Saarbrücken}
  \country{Germany}}
\affiliation{%
  \institution{Saarland University}
  \city{Saarbrücken}
  \country{Germany}}

\author{Omair Shahzad Bhatti}
\affiliation{%
  \institution{German Research Center for Artificial Intelligence (DFKI)}
  \city{Saarbrücken}
  \country{Germany}}

\author{Abdulrahman Mohamed Selim}
\affiliation{%
  \institution{German Research Center for Artificial Intelligence (DFKI)}
  \city{Saarbrücken}
  \country{Germany}}
\affiliation{%
  \institution{University of Oldenburg}
  \city{Oldenburg}
  \country{Germany}}

\author{Daniel Sonntag}
\affiliation{%
  \institution{German Research Center for Artificial Intelligence (DFKI)}
  \city{Saarbrücken}
  \country{Germany}}
\affiliation{%
  \institution{University of Oldenburg}
  \city{Oldenburg}
  \country{Germany}}

\renewcommand{\shortauthors}{Kadir et al.}

\begin{abstract}
Large language models (LLMs) used for Web information processing rely on retrieving relationships between network entities, such as domain names and IP addresses, but the internal components that causally enable this retrieval remain unclear, hindering targeted safeguards against unsafe actions. Therefore, we investigate whether their retrieval capabilities can be localized to specific attention heads or individual neurons within these LLMs. Across five LLMs from three architecture families, we identify heads through causal ablation screening, test their selectivity against matched negative and context-free controls, rank neurons within these heads using causal ablation and correlational measures, and leverage features from both causally validated heads and individual neurons to build network information retrieval detectors, which we evaluate without retraining on an independent reverse-DNS dataset. In every model, a small set of causally validated heads (1--9 out of 128--1152 candidates) builds a detector with 99.5--100\% held-out accuracy. At the neuron level, the results are model-specific, e.g., in one model, causal and correlational rankings identify the same neuron, matching full-head performance (AUC = 1.000), while in another, the best causally ranked neuron achieves only AUC = 0.665 despite the full head reaching 1.000. On independent CAIDA data, head detectors maintain AUC $\geq 0.971$, while single-neuron detectors fall below chance in three of five models. Overall, these findings highlight the importance of choosing an appropriate level of granularity when causally localizing how LLMs retrieve structured network information from context.
\end{abstract}


\begin{teaserfigure}
  \includegraphics[width=\textwidth]{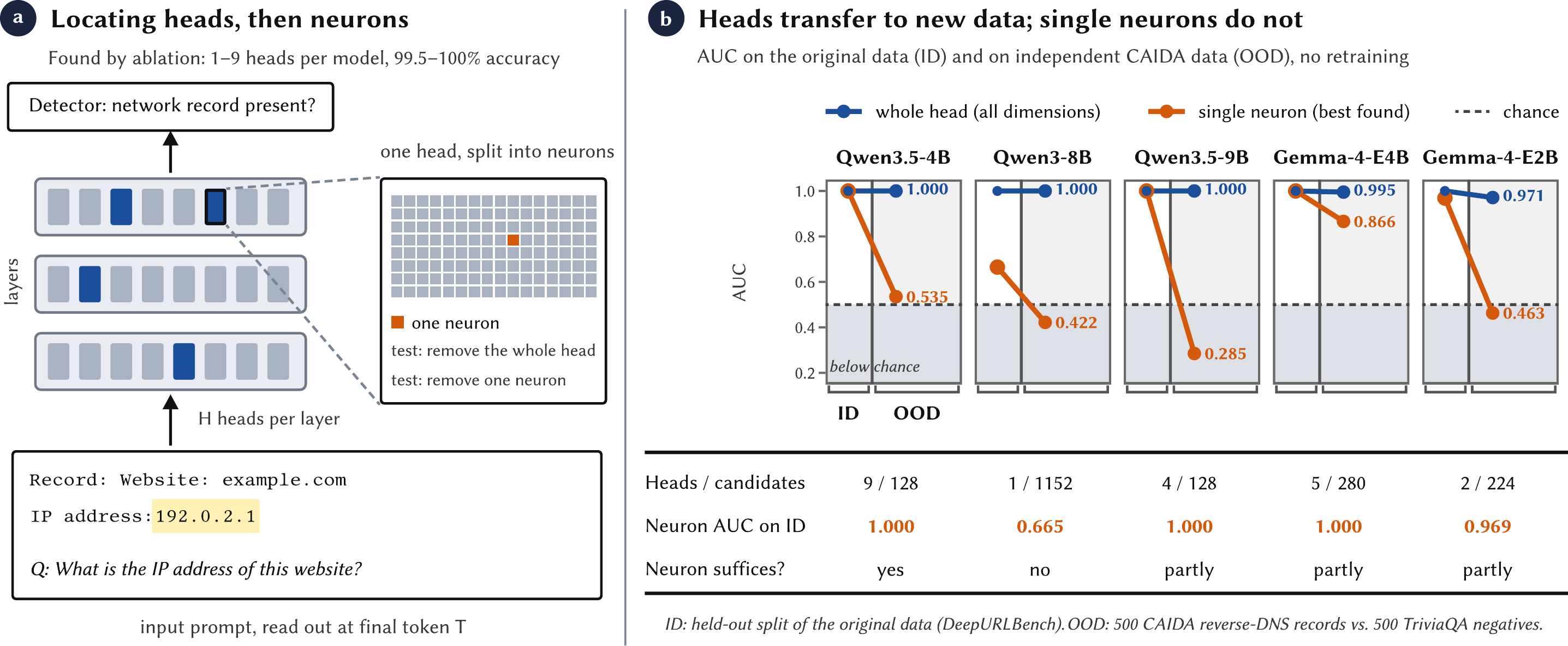}
  \caption{Overview of causal localization of structured Web
  infrastructure information retrieval.
  (a)~Given a prompt that contains a domain--IP record (an illustrative
  record is shown), we remove attention heads and keep the few that are
  causally important and selective (1--9 per model); a detector built on
  them reaches 99.5--100\% held-out accuracy. We then test whether a
  single neuron, one dimension of a head's output, carries the signal as
  well as the whole head.
  (b)~AUC of the whole-head and best single-neuron detectors on the
  original data (ID) and on independent CAIDA reverse-DNS data (OOD),
  without retraining. The whole-head detector keeps AUC $\geq 0.971$ on
  OOD data, whereas the single-neuron detector degrades and falls below
  chance (AUC $< 0.5$) in three of the five models. Whether one neuron
  suffices is model-specific.}
  \Description{Left panel: a prompt box containing an example website
  name and IP address and a question asking for the IP address. An arrow
  leads up through a stack of three layer rows, each made of eight small
  squares with a few filled blue to mark the selected attention heads,
  to a detector box asking whether a network record is present. One head
  is enlarged into a grid of small squares with a single orange square
  marking one neuron. Right panel: five small plots, one per model, each
  split by a vertical line into an in-distribution (ID) half and an
  independent-data (OOD) half. A blue line for the whole head stays near
  1.0 in every plot. An orange line for the best single neuron drops,
  ending at 0.535, 0.422, 0.285, 0.866 and 0.463, with a shaded region
  below a dashed chance line at 0.5. A table under the plots lists, for
  each model, the number of selected heads out of the candidates, the
  single-neuron AUC on ID data, and whether one neuron suffices (yes, no
  or partly).}
  \label{fig:teaser}
\end{teaserfigure}


\maketitle

\section{Introduction}
\label{sec:intro}

Mechanistic interpretability investigates how internal model components,
such as attention heads and neurons, contribute to model behavior
\citep{Feldhus_2025}. Previous studies have investigated the internal components involved
in factual knowledge using knowledge neurons
\citep{dai2022knowledge}, causal tracing to feed-forward layers
\citep{meng2022locating}, and knowledge circuits comprising
information heads, relation heads, and MLPs
\citep{yao2024knowledge}. However, which attention heads and individual neurons causally
contribute to retrieving relationships between network entities,
such as domain names and IP addresses, remains unclear.
Identifying these components is important for understanding how
Large language models (LLMs) used for Web information processing retrieve structured network
information from context. Such understanding could help researchers
develop more targeted methods for monitoring or intervening in
network information retrieval, potentially supporting safeguards
against unsafe actions.

\begin{figure*}[t]
\centering
\includegraphics[width=\textwidth]{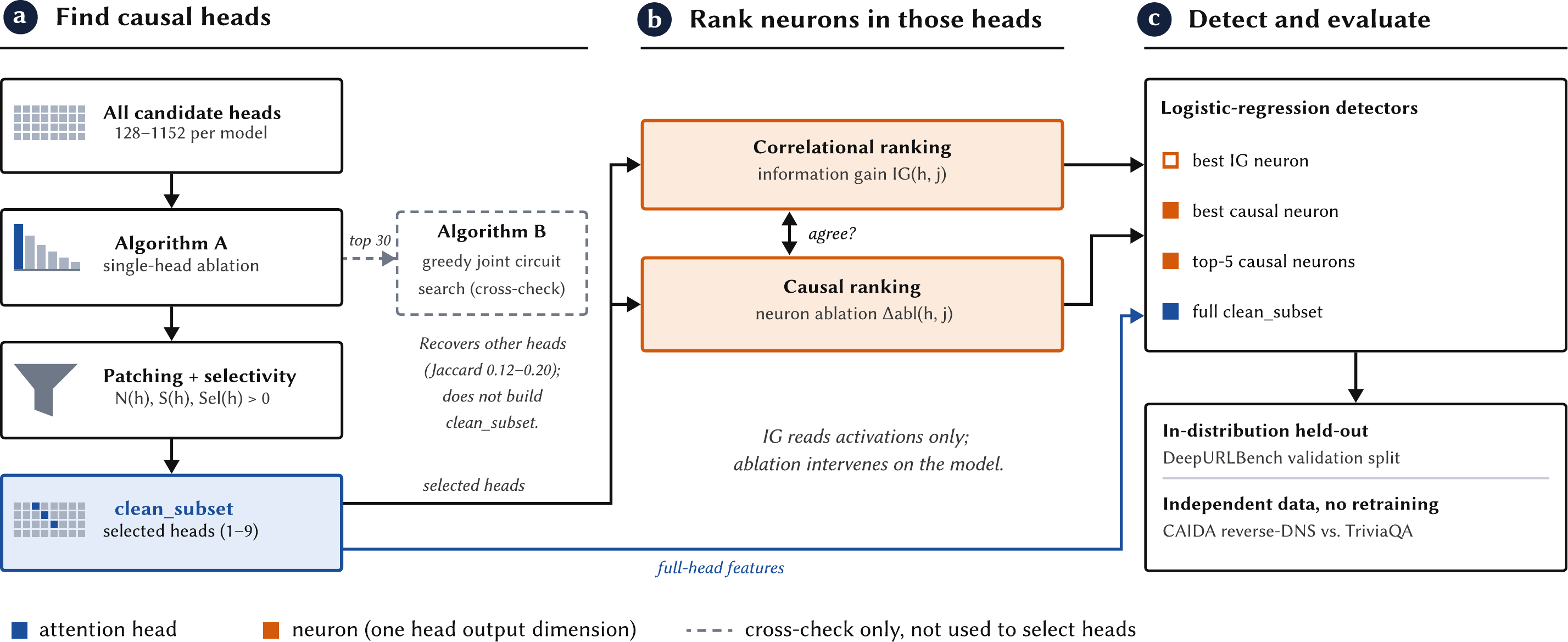}
\caption{Pipeline overview. Algorithm A screens candidate attention
heads through single-head causal ablation, followed by activation
patching and selectivity filtering to obtain the selected attention
heads (\texttt{clean\_subset}). Algorithm B performs greedy joint
circuit discovery on the top 30 heads from Algorithm A's ablation
ranking as a cross-check; it does not construct \texttt{clean\_subset}.
The selected heads are then analyzed at the neuron level with a
correlational (information-gain) and a causal (ablation) ranking.
Detectors built from single neurons, the top-5 causal neurons, or the
full selected heads are evaluated on the held-out split and, without
retraining, on independent CAIDA data.}

\Description{A three-stage flowchart read left to right. Stage 1, find
causal heads: all candidate attention heads go to Algorithm A for
single-head ablation, then to activation patching and selectivity
filtering, which produces the selected attention heads called
clean\_subset. A dashed side branch shows Algorithm B, a greedy joint
circuit search on the top 30 heads, used only as a cross-check that
does not construct clean\_subset. Stage 2, rank neurons in those heads:
clean\_subset feeds a correlational information-gain ranking and a
causal neuron-ablation ranking, with a double arrow indicating that
their top neurons are compared. Stage 3, detect and evaluate:
logistic-regression detectors use features from the best information-gain
neuron, the best causal neuron, the top-5 causal neurons, or the full
clean\_subset heads, and are evaluated on the in-distribution held-out
split and on independent CAIDA data without retraining.}

\label{fig:pipeline}
\end{figure*}

We study network information retrieval using a task in which an LLM
retrieves a domain name or IP address from a record provided in its
context. This allows us to evaluate retrieval against known ground
truth rather than treating network-relatedness as an unobserved property.
Across five LLMs from three architecture families, we identify
attention heads through causal ablation screening, evaluate their
selectivity against matched negative and context-free controls,
and compare causal and correlational rankings of individual neurons.
We then train network information retrieval detectors using features
from the selected heads and neurons and evaluate their generalization
to an independent reverse-DNS dataset without retraining.
Figure~\ref{fig:pipeline} summarizes this approach.

\textbf{Findings.} (1) At the head level, a small set of causally validated,\linebreak
selectivity-confirmed heads exists in every model tested
(Section~\ref{sec:heads}), supporting a 99.5--100\%-accurate detector on
held-out data in all five. (2) Whether that head-level responsibility is
reducible to one neuron is model-specific, not a fixed property of ``how
good the head is'' (Section~\ref{sec:neurons}): one model's top head
concentrates cleanly into a single neuron that two independent methods
agree on; another model's equally effective head is spread
across its full dimensionality. (3) The head-level result, not the
neuron-level one, is the one that transfers to an independent,
real-world dataset (Section~\ref{sec:caida}). Figure~\ref{fig:teaser} previews these findings.

\textbf{Contributions.} This work provides a systematic, causally
validated analysis of network information retrieval at both the
attention-head and individual-neuron levels across five LLMs.
By combining causal localization with neuron-level comparisons
and evaluation on an independent reverse-DNS dataset, we show
why identifying a causally important component is not sufficient
to establish that its individual neurons provide equally reliable
retrieval-related signals. These findings help distinguish the
granularity at which retrieval can be causally localized from
the granularity at which reliable detection is possible, providing
a foundation for more targeted analysis of LLMs used for Web
information processing.

\section{Related Work}
\label{sec:related}

\textbf{Causal ablation for in-context retrieval heads.} Closest to our
head-discovery setup, \citet{wu2025retrieval} identify a sparse,
largely universal set of ``retrieval heads'' in long-context models
using a retrieval-score metric and validate them causally by pruning
against a matched random-head baseline, the same logic as our random
baselines in Section~\ref{sec:neurons}. Their heads are discovered for
generic long-context copying; we ask the same causal question for one specific,
previously unlocalized entity type (network/infrastructure records) and
additionally cross-validate the discovery with a second, structurally
different algorithm (Section~\ref{sec:method}).

\textbf{Head-level specialization for facts and reasoning.}
\citet{geva2023dissecting} use interventions on attention edges to trace
factual-recall information flow and find upper-layer heads performing
the final extraction of the attribute from the subject, the closest
structural analogue, in the pure-factual-recall literature, to our task of
extracting an IP address from a hostname. \citet{yu2023characterizing}
causally scale a single head to resolve conflicts between a model's
parametric memory and new in-context information; \citet{hakimi2025timecourse}
track heads across pretraining and find they churn in and out of
entity-/relation-/fact-specific roles far more than feed-forward layers
do; \citet{zhang2026entropy} identify heads from a behavioral signature
(per-step logit entropy during chain-of-thought) rather than attention
mass directly; and \citet{li2026animacy} find 22 heads in GPT-2, split
into two causally opposing populations, mediating a single syntactic
choice. These studies find sparse, causally separable head populations for
factual, reasoning and syntactic behaviors. None of them study
network/infrastructure records, the entity type we ask about.
\citet{michel2019sixteen} show that many heads can be pruned with
little loss, and \citet{conmy2023acdc} automate circuit discovery by
ablation.

\textbf{Neuron-level factual knowledge.} \citet{dai2022knowledge}
introduce knowledge neurons, individual neurons causally tied to
specific facts; \citet{meng2022locating} causally trace factual
predictions to middle-layer MLPs and edit them directly, the precedent
for treating neuron-level causal localization as a first-class target
rather than an afterthought to head-level results. Several recent
studies complicate the single-neuron picture, which relates directly to our
finding (Section~\ref{sec:neurons}): \citet{chen2024journey} and
\citet{cao2026onemind} find \emph{language-independent} knowledge
neurons alongside \emph{degenerate} ones, i.e., several neurons
redundantly carrying the same fact, so ablating any one alone leaves
the others able to recover it; \citet{chen2025cracking} give a
structural account of these degenerate neurons and a clustering method
to find them at arbitrary size; and \citet{wang2024unveiling} show
factual recall often bypasses the knowledge-neuron pathway entirely via
a shortcut during multi-hop reasoning. This redundancy account is
the one we offer for why two independently-computed neuron
rankings can agree on \emph{which head} matters while disagreeing on
\emph{which single dimension} inside it is responsible, in three of our
five models (Section~\ref{sec:neurons}).

\textbf{Knowledge circuits and the mechanistic basis of head specialization.} \citet{yao2024knowledge} combine information heads,
relation heads, and MLPs into full knowledge circuits.
\citet{franco2026singular} give a more formal account of why attention
heads admit a feature-aligned read-out at all, via their singular
vectors, and \citet{lasnier2026translation} independently report the
same qualitative pattern we do in an unrelated task: LLM-based machine
translation decomposes into disjoint, causally separable head
populations (target-language identification vs.\ semantic-equivalence
preservation) under targeted ablation. 
\citet{chen2026mechanistic} ask which training examples caused an interpretable unit to form in the first place; this question comes before all of the above, and it is a natural next step for the heads and neurons we localize here. 
None of this circuit- and mechanism-level work addresses network/infrastructure information, at either head or neuron level, with causal validation. This is the gap we target.

\textbf{Why causal importance and decodability can diverge.}
\citet{elazar2021amnesic} causally remove a linguistic property from
sentence representations and show that probing performance does not
track the property's causal importance to model behavior. Their setting
differs from ours (sentence-level linguistic properties rather than
attention heads or neurons, and nullspace projection rather than
ablation), but the pattern is the one we report at neuron granularity in
Section~\ref{sec:neurons}: a causal intervention and a correlational
ranking agree on the responsible neuron in only one of our five models.

\textbf{Our contribution relative to this literature.} To our knowledge,
none of the above combines causal head discovery, a second cross-validating
discovery algorithm, neuron-level causal ablation (rather than a purely
correlational ranking), and validation against an independent
real-world data source, for a network/infrastructure entity specifically.
We combine all four in this paper.

\section{Method}
\label{sec:method}

Our goal is to find which attention heads, and which individual neurons
inside them, causally support retrieving a domain name or IP address from
a model's context, and to test whether these findings hold on independent
data. We first define a controlled retrieval task with control conditions,
and describe the models, splits, and evaluation. We then show how heads and neurons can be removed
individually (Section~\ref{sec:formalism}), how we discover causally
responsible heads (Section~\ref{sec:head_discovery}), and how we rank
neurons inside them (Section~\ref{sec:neuron_discovery}).

\subsection{Task and Data}

Each example presents a short context record ---
\texttt{Record: Website: <domain> / IP address: <ip>} --- followed by a
question requiring the model to copy back one specific field (e.g.\
``What is the IP address of this website?''). Because both fields are
already present in the prompt, this is a retrieval task, not a
factual-recall task: the model needs to find and copy a span from its own
context, not know anything about the real world. Domains and IP addresses
are drawn from DeepURLBench~\citep{schvartzman2024deepurlbench}, a public
dataset of URLs with DNS resolution records. Three negative conditions test whether a head's effect is specific to the
retrieval event rather than to the record or to generic
instruction-following (Table~\ref{tab:worked_example}): \emph{entity}
keeps the record but asks for its category, \emph{vocabulary} keeps the
record but asks what ``IP'' stands for, and \emph{ordinary} drops the
record and asks an arithmetic question.
A \emph{paraphrase} condition preserves the same record and answer while
rewording the retrieval question and is treated as a positive robustness
condition. We additionally use two context-free controls, \emph{question\_only} (the
retrieval question without the record) and \emph{vocab\_no\_context} (the
vocabulary question without the record), to test for false positives
driven by wording alone. In the appendices the datasets of these conditions are
written \emph{D\_network}, \emph{D\_paraphrase}, \emph{D\_entity},
\emph{D\_vocab} and \emph{D\_ordinary}.

\subsection{Intervening on Attention Heads and Neurons}
\label{sec:formalism}

Multi-head attention \citep{vaswani2017attention} sums the contributions of
its heads (Eq.~\eqref{eq:decomp}), so a single head, or one output dimension
within it (which we call a neuron), can be removed on its own. Let
$x\in\mathbb{R}^{T\times d}$ denote a layer's input hidden states, where
$T$ is the number of tokens and $d$ is the residual-stream width. Let
$H$ denote the number of query attention heads and let $d_h$ denote the
model-defined head dimension; importantly, we do not assume
$d_h=d/H$.

For query head $h\in\{1,\dots,H\}$,
\begin{equation}
Q_h = xW_Q^{(h)},\quad K_h = xW_K^{(h)},\quad V_h = xW_V^{(h)},
\label{eq:qkv}
\end{equation}
where $K_h$ and $V_h$ denote the key/value projections used by query
head $h$. In grouped-query attention \citep{ainslie2023gqa}, multiple query heads may share
the same key/value head. The corresponding attention output is
\begin{equation}
A_h = \mathrm{softmax}\!\left(\frac{Q_hK_h^\top}{\sqrt{d_h}}\right),
\qquad O_h = A_hV_h \in \mathbb{R}^{T\times d_h}.
\label{eq:attn}
\end{equation}

Concatenating the $H$ query-head outputs gives a representation of width
$Hd_h$, which is projected back to the residual-stream width by
$W_O\in\mathbb{R}^{Hd_h\times d}$. Writing
$W_O^{(h)}\in\mathbb{R}^{d_h\times d}$ for the block corresponding to
head $h$,
\begin{align}
\mathrm{Attn}(x)
&= \big[O_1;\dots;O_H\big]W_O
 = \sum_{h=1}^{H} O_h W_O^{(h)}
\notag\\
&= \sum_{h=1}^{H}\sum_{j=1}^{d_h}
O_h[:,j]\,W_O^{(h)}[j,:].
\label{eq:decomp}
\end{align}

\begin{figure*}[t]
\centering
\includegraphics[width=\textwidth]{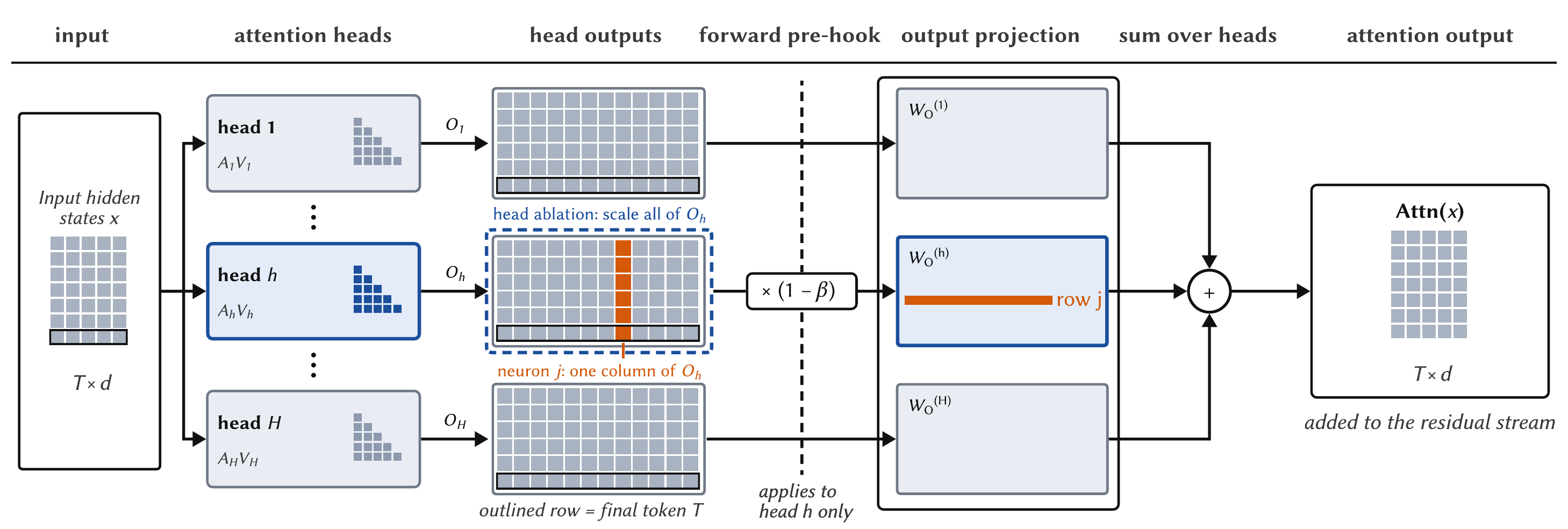}
\caption{Head-level and neuron-level interventions in an attention
layer. Head ablation (blue) suppresses the entire head output $O_h$,
whereas neuron ablation (orange) suppresses one output dimension
$O_h[:,j]$ and its corresponding projected contribution
$O_h[:,j]W_O^{(h)}[j,:]$ (Eq.~\eqref{eq:decomp}).
Both interventions operate across all sequence positions through
a forward pre-hook, with full ablation at $\beta=1$.
Activation patching and feature extraction use only the final
token position $T$ (outlined row).}
\Description{An input matrix feeds multiple attention heads, each
producing an output matrix. Head h is highlighted in blue, with
a dashed blue outline indicating whole-head ablation and an
orange column indicating neuron j. A forward pre-hook scales
the selected component before output projection. The corresponding
projection block contains an orange row representing the weights
associated with neuron j. Projected head outputs are summed to
produce the attention output. The outlined row in each head output
marks the final token position used for activation patching and
feature extraction.}
\label{fig:mechanism}
\end{figure*}

All interventions in this paper follow from Eq.~\eqref{eq:decomp}: zeroing
$O_h$ removes head $h$'s whole term from the sum without touching any other
head; zeroing one column $O_h[:,j]$ removes only that one additive term inside head $h$'s own
contribution.
Figure~\ref{fig:mechanism} shows where each intervention sits in this
computation. Ablation (Sections~\ref{sec:head_discovery}
and~\ref{sec:neuron_discovery}) scales $O_h$ or $O_h[:,j]$ at \emph{every}
sequence position in the forward pass, so a suppressed head's influence
is blocked not only at the final position but also along any path that
re-enters the residual stream at an earlier position and is read again by
a later layer; necessity/sufficiency patching and every captured feature
vector, by contrast, are restricted to the single final position $t=T$,
where the answer is produced, since those operations read or inject one
specific vector rather than silencing a component outright.

\subsection{Causal Head Discovery}
\label{sec:head_discovery}

For each candidate head, HeadSuppressor (Algorithm A in
Figure~\ref{fig:pipeline}) replaces $O_h$ with $(1-\beta)O_h$ at every
position for $\beta\in[0,1]$ ($\beta=1$: full removal), as in head-pruning
studies \citep{michel2019sixteen}, and we measure the resulting drop in the
log-probability of the correct answer's first token $y^*$, read out at the
final position:
\begin{equation}
\Delta_{\mathrm{abl}}(h) = \mathbb{E}_{x}\Big[\log p_\theta(y^*\!\mid\! x)
 - \log p_{\theta}(y^*\!\mid\! x;\, O_h\!\leftarrow\!0)\Big],
\label{eq:dabl}
\end{equation}
a causal-effect ranking that does not depend on attention weight at all.
We then run activation patching \citep{zhang2024patching} on the top 50 screened heads. Let
$x^{(E)}$ be the retrieval prompt and $x^{(C)}$ the entity-control
prompt sharing the same context prefix; necessity asks whether removing
a head's own output hurts the retrieval answer, and sufficiency asks
whether injecting that head's retrieval-condition output into the
control prompt raises the retrieval answer's probability anyway, although
the control question does not ask for it:
\begin{align}
N(h) &= \log p_\theta(y^*\!\mid\! x^{(E)}) - \log p_\theta\big(y^*\!\mid\! x^{(E)};\,O_h\!\leftarrow\!O_h^{(C)}\big),\label{eq:nec}\\
S(h) &= \log p_\theta\big(y^*\!\mid\! x^{(C)};\,O_h\!\leftarrow\!O_h^{(E)}\big) - \log p_\theta(y^*\!\mid\! x^{(C)}).
\label{eq:suff}
\end{align}
Finally, we filter to heads with positive selectivity --- ablating a head
must hurt the target retrieval task more than it hurts the three matched
negative conditions. With $\Delta_{\mathrm{abl}}^{\mathrm{net}},
\Delta_{\mathrm{abl}}^{\mathrm{ent}}, \Delta_{\mathrm{abl}}^{\mathrm{voc}},
\Delta_{\mathrm{abl}}^{\mathrm{ord}}$ the ablation effect
(Eq.~\eqref{eq:dabl}) on the network, entity, vocabulary, and ordinary
datasets respectively,
\begin{equation}
\mathrm{Sel}(h) = \Delta_{\mathrm{abl}}^{\mathrm{net}}(h)
 - \tfrac13\big(\Delta_{\mathrm{abl}}^{\mathrm{ent}}(h)+\Delta_{\mathrm{abl}}^{\mathrm{voc}}(h)+\Delta_{\mathrm{abl}}^{\mathrm{ord}}(h)\big),
\label{eq:sel}
\end{equation}
and the top 20 by combined necessity and sufficiency, among heads with
positive selectivity, form the discovered head set $H_E$:
\begin{equation}
H_E = \operatorname*{top\text{-}20}_{h\,:\,\mathrm{Sel}(h)>0}\ \big(N(h)+S(h)\big).
\label{eq:HE}
\end{equation}

To check that $H_E$ is not an artifact of this one screening procedure,
we additionally run a second, structurally different algorithm
(Algorithm B in Figure~\ref{fig:pipeline}): greedy joint circuit discovery
\citep{conmy2023acdc}, which builds a \emph{set} of heads by greedily
suppressing a growing group together rather than one head at a time,
picking at each step whichever remaining head produces the largest
additional drop given the heads already suppressed. With candidate pool
$\mathcal{C}$ (the top 30 heads by $\Delta_{\mathrm{abl}}$) and the joint
ablation loss of a head set $S$,
$\mathcal{L}(S)=\mathbb{E}_x[\log p_\theta(y^*\!\mid\! x)-\log p_\theta(y^*\!\mid\! x;\{O_h\!\leftarrow\!0\}_{h\in S})]$,
\begin{equation}
S_0=\emptyset,\quad
h_t^\star = \operatorname*{arg\,max}_{h\in\mathcal{C}\setminus S_{t-1}}
 \big[\mathcal{L}(S_{t-1}\cup\{h\}) - \mathcal{L}(S_{t-1})\big],
\label{eq:greedy}
\end{equation}
with $S_t=S_{t-1}\cup\{h_t^\star\}$, stopping once a step's marginal gain
falls below 5\% of $\mathcal{L}(S_1)$ or after 10 heads. Unlike Eq.~\eqref{eq:dabl}, this
measures each head's contribution given every previously-selected head
already suppressed, so it can credit a head that independent screening
misses due to redundancy between heads. Section~\ref{sec:ablation} reports
how far the head set recovered by Algorithm B overlaps $H_E$.

A causally important head can still fail a basic specificity check ---
for example, by also firing on contexts that merely mention network
vocabulary without containing a real record. For every candidate head we
fit a linear probe \citep{hewitt2019control} on its activation and test it against a
vocabulary-only hard negative; heads that flag this at a high rate are
discarded. The surviving heads, \texttt{clean\_subset}, are used for
every neuron-level result below. Section~\ref{sec:ablation} compares a
detector trained on the unfiltered $H_E$ with one trained on
\texttt{clean\_subset}.

\subsection{Causal Neuron Discovery}
\label{sec:neuron_discovery}

For head $h\in\texttt{clean\_subset}$ and neuron $j\in\{1,\dots,d_h\}$,
\texttt{Neuron\allowbreak Suppressor} replaces
$O_h[:,j]$ with $(1-\beta)O_h[:,j]$ at every position, giving the
neuron-level analogue of Eq.~\eqref{eq:dabl}, an intervention rather
than a correlational read of the activation:
\begin{equation}
\Delta_{\mathrm{abl}}(h,j) = \mathbb{E}_x\big[\log p_\theta(y^*\!\mid\! x) - \log p_\theta(y^*\!\mid\! x;\,O_h[:,j]\!\leftarrow\!0)\big].
\label{eq:neuron_ablation}
\end{equation}
We compare this against a purely correlational ranking used elsewhere in
the knowledge-neuron literature: quantile-bin each neuron's activation
and measure how much knowing the bin reduces label entropy, with no
model intervention at all. With label $Y\in\{0,1\}$ (network vs.\
ordinary) and $B_{h,j}\in\{1,\dots,8\}$ the quantile bin of $O_h[T,j]$
over the calibration set,
\begin{equation}
\mathrm{IG}(h,j) = H(Y) - \sum_{b} P(B_{h,j}\!=\!b)\, H(Y\mid B_{h,j}\!=\!b),
\label{eq:ig}
\end{equation}
with $H(\cdot)$ binary entropy. Eq.~\eqref{eq:ig} never touches
$p_\theta$, which is why it can disagree with
Eq.~\eqref{eq:neuron_ablation} (Table~\ref{tab:neuron_agree}): the two
equations measure different properties of the same scalar.

We then build four classifiers per model, trained on
\{network, paraphrase\} positives vs.\ \{ordinary, vocabulary, entity\}
negatives --- from the single best causal neuron, the top-5 causal
neurons, the single best information-gain neuron, and the full
\texttt{clean\_subset} --- all using the same logistic regression over a
feature vector $z$, either one neuron's scalar $O_h[T,j]$ or a
concatenation of full head vectors $O_h[T,:]$:
\begin{equation}
\hat p(y\!=\!1\mid z) = \sigma(w^\top z + b).
\label{eq:clf}
\end{equation}

\section{Experiments and Results}
\label{sec:experiments}

\textbf{Setup.} We write $P(\text{internet})$ for the detector's predicted
probability that the context contains network information; a prompt is
\emph{flagged} when $P\geq 0.5$. We use the instruction-tuned checkpoints
\texttt{Qwen/Qwen3.5-4B}, \texttt{Qwen/Qwen3.5-9B}, \texttt{Qwen/Qwen3-8B},
\texttt{google/\allowbreak gemma-\allowbreak 4-\allowbreak E2B-\allowbreak it} and
\texttt{google/\allowbreak gemma-\allowbreak 4-\allowbreak E4B-\allowbreak it}
(the Qwen3 family is described in \citealp{yang2025qwen3}). Head discovery (Section~\ref{sec:head_discovery}) uses a calibration split
of 200 items per condition; the greedy circuit search and the neuron
ablation screen (Sections~\ref{sec:head_discovery}
and~\ref{sec:neuron_discovery}) use 120 items from the validation split,
and classifier accuracies are computed on the validation split.
Every detector is a scikit-learn~\citep{scikit-learn} logistic regression ($C=1$, balanced class
weights) on the captured activations, with positives
\{network, paraphrase\} and negatives \{ordinary, vocabulary, entity\}. All results come from a single run per model. Where a table reports a
mean $\pm$ standard deviation, that spread is over 20 draws of a random
neuron (Table~\ref{tab:neuron_random_control}), not over repeated runs of
head discovery.

\textbf{Worked example.} Table~\ref{tab:worked_example} shows one held-out
record under every condition, with the output probability of the fixed
Qwen3.5-4B detector. The \emph{network}, \emph{paraphrase}, \emph{entity} and
\emph{vocabulary} conditions share one DeepURLBench record
(\texttt{al-mostafa.com}, \texttt{62.\allowbreak 151.\allowbreak 181.\allowbreak 154}, labeled benign);
\emph{ordinary}, \emph{question\_only} and \emph{vocab\_no\_context} have an
empty context. The detector fires only when the context contains the record and
the question asks for one of its fields (\emph{network} and \emph{paraphrase},
$P\!=\!1.000$ for both). It stays silent on \emph{entity}, which has the same
context but asks for the category ($P\!=\!0.000$), and on \emph{question\_only},
which asks the same retrieval question without a record ($P\!=\!0.024$; this
wording-only residual is discussed in Section~\ref{sec:discussion}). Appendix~\ref{app:worked_examples} repeats the example for both retrieval
directions, on two records, across all five models.

\begin{table*}[t]
\caption{A worked example: one held-out record under every condition, with
the predicted probability from the fixed Qwen3.5-4B detector. The
\emph{network}, \emph{paraphrase}, \emph{entity} and \emph{vocabulary} rows
share the same context (\texttt{al-mostafa.com}, \texttt{62.151.181.154});
the other rows have an empty context.}
\label{tab:worked_example}
\centering
\small
\begin{tabular}{llp{5.4cm}lcc}
\toprule
Condition & Context & Question & True answer & $P(\text{internet})$ & Flagged \\
\midrule
network & record & What is the IP address of this website? & 62.151.181.154 & 1.000 & yes \\
paraphrase & record & Resolve the IPv4 address associated with this domain. & 62.151.181.154 & 1.000 & yes \\
entity & record & What is the category of this website? Answer with one word: benign, mal, or phishing. & benign & 0.000 & no \\
vocabulary & record & In one short phrase, what does the abbreviation ``IP'' stand for? & Internet Protocol & 0.000 & no \\
ordinary & empty & What is 67 + 14? & 81 & 0.001 & no \\
question\_only & empty & What is the IP address of this website? & 62.151.181.154 & 0.024 & no \\
vocab\_no\_context & empty & In one short phrase, what does the abbreviation ``IP'' stand for? & Internet Protocol & 0.001 & no \\
\bottomrule
\end{tabular}
\end{table*}

\subsection{A Clean Head Set Exists in Every Model}
\label{sec:heads}

\begin{table}[t]
\caption{A causally-validated, selectivity-confirmed head set exists in
every model tested, regardless of architecture or candidate-pool size.}
\label{tab:heads}
\centering
\small
\begin{tabular}{lrrrr}
\toprule
Model & Cand. & Clean & Acc. & AUC \\
 & heads & heads & (held-out) & \\
\midrule
Qwen3.5-4B   & 128  & 9 & 100.0\% & 1.000 \\
Qwen3-8B     & 1152 & 1 &  99.5\% & 1.000 \\
Qwen3.5-9B   & 128  & 4 & 100.0\% & 1.000 \\
Gemma-4-E4B  & 280  & 5 &  99.9\% & 1.000 \\
Gemma-4-E2B  & 224  & 2 &  99.9\% & 1.000 \\
\bottomrule
\end{tabular}
\end{table}

In all five models, a set of one to nine causally validated heads supports a detector with
99.5--100\% held-out accuracy and AUC 1.000 on data from the source used for
discovery (Table~\ref{tab:heads}). This holds across attention designs (hybrid
linear/full attention in Qwen3.5, dense attention in Qwen3-8B, interleaved
sliding-window/global attention in Gemma-4), across candidate pools of 128 to
1152 heads, and even when only one head survives the selectivity filter
(Qwen3-8B). The following subsections report an ablation of this procedure
(Section~\ref{sec:ablation}), how far this localization narrows to
individual neurons (Section~\ref{sec:neurons}), and whether it holds on data
the heads were not discovered on (Section~\ref{sec:caida}).

\subsection{Ablations of the Head-Finding Procedure}
\label{sec:ablation}

Algorithm B recovers, in every model, a head set that overlaps only partly
with $H_E$ (Jaccard 0.12--0.20, i.e.\ 3--5 of 20 heads) and is equally
damaging. The underlying causal signal therefore does not depend on which
algorithm finds it, even though the two algorithms name different heads as
most responsible.

\textbf{Does the filter matter?} $H_E$ already reaches near-perfect held-out
accuracy and AUC, so the filter might seem unnecessary. To test this, we train
the same detector on the unfiltered 20 heads of $H_E$ instead of
\texttt{clean\_subset} and compare both on the held-out split and on the
context-free \texttt{question\_only} control (Table~\ref{tab:stage_ablation}):

\begin{table}[t]
\caption{Unfiltered $H_E$ vs.\ filtered \texttt{clean\_subset}: held-out
AUC is identical either way, but the context-free wording confound
(\texttt{question\_only} flag rate) differs: the filter lowers it to 0 only
in the Qwen models.}
\label{tab:stage_ablation}
\centering
\small
\setlength{\tabcolsep}{5pt}
\begin{tabular}{lccc}
\toprule
Model & AUC & \multicolumn{2}{c}{\texttt{question\_only} flag rate} \\
\cmidrule(lr){3-4}
 & (both) & $H_E$ (20) & \texttt{clean\_subset} \\
\midrule
Qwen3.5-4B   & 1.000 & 1.00 & \textbf{0.00} \\
Qwen3-8B     & 1.000 & 1.00 & \textbf{0.00} \\
Qwen3.5-9B   & 1.000 & 1.00 & \textbf{0.00} \\
Gemma-4-E4B  & 1.000 & 1.00 & 1.00 \\
Gemma-4-E2B  & 1.000 & 1.00 & 1.00 \\
\bottomrule
\end{tabular}
\end{table}

Held-out AUC is 1.000 with or without the filter, and accuracy differs by at
most 0.3 points. The unfiltered $H_E$ detector, however, flags 100\% of
context-free retrieval questions in all five models, so it responds to the
wording of the question. The filtered \texttt{clean\_subset} removes this in
the three Qwen models (0\%) but not in either Gemma model (still
100\%). Held-out accuracy and AUC cannot detect this failure; only the
context-free control does. The filter is therefore not redundant: it leaves the
headline metrics unchanged but changes how the detector behaves on a control
that those metrics miss. In the Gemma models it does not remove the confound, so
their high held-out scores should not be read as showing that the detector
ignores the wording of the question. The same Qwen-versus-Gemma difference appears in
the negative-class errors of Section~\ref{sec:caida}. A same-sized random-head control is reported in
Appendix~\ref{app:causal_dns_mac}. Sufficiency is higher for
\texttt{clean\_subset} than for the random set in every model and every
condition, including the original network condition
(Table~\ref{tab:causal_dns_mac}).

\subsection{Does Responsibility Concentrate Into a Neuron?}
\label{sec:neurons}

Section~\ref{sec:neuron_discovery} gives, per model, two independent
rankings of the neurons inside \texttt{clean\_subset} and four classifiers
built from them. Table~\ref{tab:neuron_acc} shows how much of the head's
detection performance a single neuron recovers, and
Table~\ref{tab:neuron_agree} whether the two rankings agree.

\begin{table}[t]
\caption{Single-neuron (best causally ranked neuron) vs.\ full-head
detection performance (held-out accuracy at the 0.5 threshold, and AUC).}
\label{tab:neuron_acc}
\centering
\small
\begin{tabular}{lrrrr}
\toprule
 & \multicolumn{2}{c}{Single neuron} & \multicolumn{2}{c}{Full head} \\
Model & Acc. & AUC & Acc. & AUC \\
\midrule
Qwen3.5-4B   & 99.4\% & 1.000 & 100.0\% & 1.000 \\
Qwen3-8B     & 53.9\% & 0.665 &  99.5\% & 1.000 \\
Qwen3.5-9B   & 99.4\% & 1.000 & 100.0\% & 1.000 \\
Gemma-4-E4B  & 99.9\% & 1.000 &  99.9\% & 1.000 \\
Gemma-4-E2B  & 86.6\% & 0.969 &  99.9\% & 1.000 \\
\bottomrule
\end{tabular}
\end{table}

\begin{table}[t]
\caption{Top-ranked neuron (layer, head, dimension) under the causal
ranking and under the information-gain (IG) ranking.}
\label{tab:neuron_agree}
\centering
\small
\begin{tabular}{lll}
\toprule
Model & Causal top / IG top & Agreement \\
\midrule
Qwen3.5-4B  & (15,1,0)/(15,1,0)     & \textbf{exact match} \\
Qwen3-8B    & (29,12,82)/(29,12,27) & same head, diff.\ dim \\
Qwen3.5-9B  & (23,15,8)/(19,4,1)    & \textbf{different head} \\
Gemma-4-E4B & (32,2,141)/(32,2,7)   & same head, diff.\ dim \\
Gemma-4-E2B & (30,0,183)/(30,0,48)  & same head, diff.\ dim \\
\bottomrule
\end{tabular}
\end{table}

\textbf{Neuron-level random control.} For each model we compare the AUC of
the \texttt{single\_best\_causal\_neuron} classifier against two
random-neuron baselines, 20 repeats each: a random dimension of the
\emph{same} head, and a random (layer, head, dimension) triple from the
entire candidate pool (Table~\ref{tab:neuron_random_control}).

\begin{table}[t]
\caption{AUC of the actual \texttt{single\_best\_causal\_neuron} classifier
vs.\ two random-neuron baselines (mean $\pm$ std over 20 repeats each).}
\label{tab:neuron_random_control}
\centering
\scriptsize
\setlength{\tabcolsep}{3pt}
\begin{tabular}{lrrr}
\toprule
Model & Actual & Random (same head) & Random (any neuron) \\
\midrule
Qwen3.5-4B   & \textbf{1.000} & 0.727 $\pm$ 0.155 & 0.667 $\pm$ 0.186 \\
Qwen3-8B     & 0.665          & 0.633 $\pm$ 0.148 & 0.728 $\pm$ 0.179 \\
Qwen3.5-9B   & \textbf{1.000} & 0.680 $\pm$ 0.128 & 0.654 $\pm$ 0.195 \\
Gemma-4-E4B  & \textbf{1.000} & 0.748 $\pm$ 0.180 & 0.744 $\pm$ 0.195 \\
Gemma-4-E2B  & \textbf{0.969} & 0.675 $\pm$ 0.149 & 0.724 $\pm$ 0.188 \\
\bottomrule
\end{tabular}
\end{table}

Four of five models show the pattern the ``concentrated'' hypothesis
predicts: the actual chosen neuron (AUC $\geq 0.969$) sits above both
random-neuron baselines (AUC $0.65$--$0.75$; $1.3$--$2.5$ standard
deviations below the actual value in every case), so this neuron, and not
an arbitrary one, carries the signal. Qwen3-8B is the exception: its actual
best neuron (AUC $0.665$) falls \emph{inside} the fully-random baseline's
range ($0.728 \pm 0.179$) and lies within one standard deviation of the
within-head random baseline ($0.633 \pm 0.148$). For this model, the best
neuron the causal-ablation search could find performs no better than a
randomly chosen one, which, beyond the single-point comparison in Table~\ref{tab:neuron_acc}, supports the view that the responsibility here is not reducible to any one dimension.

In Qwen3.5-4B, a single neuron reaches the same in-distribution AUC (1.000)
as the full 9-head detector, and both rankings, computed independently, agree
on which neuron it is: responsibility here is concentrated and confirmed by
both rankings. In Qwen3-8B, the opposite holds: the one clean head works as a
whole (AUC = 1.000), but the best causally ranked neuron inside it reaches
only AUC = 0.665 (53.9\% accuracy, barely better than chance). Whatever this
head represents needs most of its 128 dimensions acting together, and the two
neuron rankings do not agree on which dimension to try. The remaining three
models fall in between: in all three, a single neuron gets close to or
matches the full head's AUC (1.000, 1.000 and 0.969), but the two rankings
disagree on the \emph{exact} neuron. In Gemma-4-E4B and Gemma-4-E2B they
still agree on the head, while in Qwen3.5-9B they select different heads.
This is consistent with several neurons carrying overlapping versions of the
same signal, rather than one privileged neuron.

Head-level localization succeeded in every model. What varies from model to
model is whether it can be pushed down to a single neuron, and this variation
is not predictable from how well the head itself performs: Qwen3-8B's head is
as good as Qwen3.5-4B's (AUC = 1.000 for both), but their neuron-level
results are opposite.

\subsection{Robustness to an Independent Data Source}
\label{sec:caida}

Every result above uses examples derived from DeepURLBench's website crawl.
This subsection does not discover new heads. It freezes each model's
detector and scores a new set of prompts. The 500 positives are reverse-DNS
records from CAIDA's IPv4 Routed /24 DNS Names dataset
\citep{caida_dns_names}: an IP address and the hostname that resolved to it,
collected by the CAIDA Ark traceroute infrastructure. These are mostly
residential-ISP and infrastructure hostnames, not the consumer websites in
DeepURLBench. The 500 negatives are held-out TriviaQA questions
\citep{joshi2017triviaqa}, with no network record in the prompt. Both sets
use the same prompt template as the rest of the paper, and the detector is
applied with no retraining. Table~\ref{tab:caida} reports accuracy on each
class and AUC for the full-head detector and for the best single-neuron
detector.

\begin{table}[t]
\caption{Cross-validation on an independent, real-world dataset (CAIDA).
The full-head detector transfers with high AUC; the single-neuron
detector does not. Pos.\ and Neg.\ are the accuracies on the CAIDA records
and on the trivia negatives.}
\label{tab:caida}
\centering
\scriptsize
\setlength{\tabcolsep}{3pt}
\begin{tabular}{lrrrrr}
\toprule
Model & Full head & Pos. & Neg. & Full head & Single neuron \\
 & acc.\ & acc. & acc. & AUC & AUC \\
\midrule
Qwen3.5-4B   & 99.9\%  & 100.0\% & 99.8\%  & 1.000 & 0.535 \\
Qwen3-8B     & 99.8\%  & 100.0\% & 99.6\%  & 1.000 & 0.422 \\
Qwen3.5-9B   & 100.0\% & 100.0\% & 100.0\% & 1.000 & \textbf{0.285} \\
Gemma-4-E4B  & 91.9\%  & 100.0\% & 83.8\%  & 0.995 & 0.866 \\
Gemma-4-E2B  & 91.5\%  & 100.0\% & 83.0\%  & 0.971 & 0.463 \\
\bottomrule
\end{tabular}
\end{table}

Every full-head detector classifies all 500 CAIDA records correctly without
retraining, so the head-level finding transfers to data the heads never saw,
collected by a different institution with a different methodology. The two
models with lower overall accuracy (Gemma-4-E4B, Gemma-4-E2B) lose points only
on the negative class: 16--17\% of the trivia questions are flagged, which
fits the wording sensitivity in Table~\ref{tab:stage_ablation}.

The single-neuron detectors transfer less well, and one (Qwen3.5-9B,
AUC = 0.285) scores \emph{below chance}: its score is anti-correlated with the
true label on this new data source, despite reaching AUC = 1.000 on the
original held-out split. Neuron-level concentration that looks clean
in-distribution is therefore not a safe substitute for the full head when the
goal is out-of-distribution robustness. Appendix~\ref{app:dns_mac} applies this same frozen detector to DNS-TTL and
MAC-address questions, which were never used in discovery or training, and
Appendix~\ref{app:causal_dns_mac} tests whether \texttt{clean\_subset} is
causally involved in retrieving those fields, against a same-sized
random-head control.
\section{Discussion and Limitations}
\label{sec:discussion}

We investigated whether network information retrieval in LLMs can be
causally localized to specific attention heads or individual neurons.
Across five models, we identified small sets of causally validated
heads and examined whether individual neurons within them provide
comparable detection performance. The results raise a broader
question about the granularity at which causal localization and
reliable detection of retrieval-related information coincide.

\textbf{Head-level versus neuron-level generalization.}
The selected attention heads support detectors that generalize to
independent CAIDA data, whereas single-neuron detectors show
substantially weaker transfer. One possible explanation is that
retrieval-related information is distributed across multiple
dimensions of a head's output. A detector using the full head can
combine information from these dimensions, while a detector relying
on one neuron may depend on a narrower feature whose relationship
to retrieval changes across data sources. This interpretation is
consistent with the model-specific neuron-level results, particularly
Qwen3-8B, where the full-head detector performs well despite the
limited detection performance of its best causally ranked neuron.
However, our experiments do not directly establish how information
is distributed across dimensions or whether combining them causes
the improved generalization.

\textbf{Implications for neuron-level interpretability.}
The disagreement between causal and correlational neuron rankings
shows why these approaches should not be treated as interchangeable.
Causal ablation measures how suppressing a neuron affects the
model's retrieval behavior, whereas information gain measures the
association between its activation and the task label. A neuron
can therefore be important under one criterion without being
equally informative under the other. Even when both rankings
identify the same neuron and its detector performs well on
held-out data, this does not guarantee generalization to an
independent data source. Our findings suggest that neuron-level
localization should be evaluated using both causal interventions
and independent detector evaluation, rather than relying on
in-distribution performance alone. This distinction is consistent
with earlier observations that causal importance and linear
decodability can diverge \citep{elazar2021amnesic}.

\textbf{Sensitivity to question wording.}
The selectivity filter reveals another limitation that is not
apparent from held-out accuracy or AUC. It eliminates false
positives on context-free retrieval questions in the three Qwen
models, but the two Gemma models continue to flag these questions.
The Gemma detectors also produce more false positives on the
independent TriviaQA negatives. One possible explanation is that
their selected head activations encode features associated with
retrieval-question wording alongside information about the
record in context. The current experiments establish this
difference in detector behavior but do not determine whether it
arises from architectural differences, model training, or the
features selected by the linear classifiers. These results
underscore the importance of negative and context-free
controls when evaluating network information retrieval detectors.

\textbf{Generalization beyond domain--IP retrieval.}
The additional DNS-TTL and MAC-address experiments help distinguish
causal involvement from detector generalization. Joint
necessity/sufficiency patching provides evidence that the selected
heads contribute to retrieving both fields across the five models
(Appendix~\ref{app:causal_dns_mac}). However, detectors trained
only on domain--IP examples identify these new retrieval conditions
less consistently, particularly for DNS-TTL
(Appendix~\ref{app:dns_mac}). This suggests that causal involvement
in related retrieval tasks does not necessarily imply that a fixed
linear decision boundary will recognize them. Consequently,
generalization should be evaluated separately for the underlying
components and for detectors trained on their activations.

\textbf{Implications for Web information processing.}
Identifying attention heads involved in network information
retrieval could support more targeted analysis of Web-facing LLM
systems. For example, detectors built from these heads could serve
as internal monitoring signals when an LLM retrieves network
information from its context before taking an action. Appendix~\ref{app:use_case} provides an initial test of this application using naturalistic
prompts outside the training template. The detectors flag all
25 positive model--item evaluations, but correctly reject only
12 of the 25 negative evaluations. These results demonstrate
the potential of head-level detectors for monitoring network
information retrieval, while highlighting the need for improved
specificity across different question types. In particular, false
positives on unrelated retrieval questions indicate that a
detector may respond to retrieval-related features more generally
rather than exclusively to network information. Further
evaluation would be necessary before such monitoring could
support reliable safeguards against unsafe actions.

\textbf{Limitations and future directions.}
Our analysis focuses on domain--IP retrieval across five LLMs
from three architecture families. Although the results reveal
model-specific differences in neuron-level localization,
further experiments are needed to determine which architectural
or representational factors explain these differences.
In particular, joint and sequential neuron ablations could
help distinguish distributed representations from redundancy
among individual neurons. While our evaluation includes
independent CAIDA data and additional network entity types,
broader testing across retrieval tasks, prompt formats, and
models would help establish how widely these results generalize.

Overall, our study highlights the importance of evaluating
causal localization and detector generalization together
at different levels of granularity, rather than relying on
in-distribution performance alone.

\vspace{-5mm}
\section{Conclusion}
\label{sec:conclusion}

Across five models and three architecture families, a small set of
causally validated attention heads identifies when a model's context contains
network information, and detectors built from these heads generalize to an
independent reverse-DNS dataset without retraining. Whether the responsibility narrows further to a single
neuron depends on the model: it concentrates in one neuron, confirmed by both
rankings, in one model; it stays spread across the head in another; and it is
partial in the remaining three. Unlike the head-level result, the neuron-level
result does not reliably transfer to the independent dataset. Causal head
discovery for network information worked in every model we tested, but a
single-neuron claim needs to be verified separately in each model. Future work
could test whether the same heads and neurons carry other network entity types,
beyond the DNS-TTL and MAC fields examined in the appendices.

\section*{Impact Statement}

This paper presents work whose goal is to advance the field of
mechanistic interpretability for language models. The task studied
(retrieving a domain name or IP address already present in a model's own
context) is a controlled probe of model internals, not
a system for extracting or inferring real individuals' or organizations'
private network information from a deployed model; all data used (a
URL/DNS dataset and reverse-DNS measurements) is already publicly
available. A possible application of retrieval detectors of this
kind is an internal monitoring signal for LLM agents, for example flagging
when a model is about to act on network/infrastructure information
retrieved from its context. Appendix~\ref{app:use_case}
provides a small proof-of-concept stress test of this idea on
naturalistic text, but we do not build or evaluate a deployment-ready
monitoring system. We are not aware of a specific negative
societal consequence of this work beyond those generally associated with
advancing interpretability methods for language models.


\bibliographystyle{ACM-Reference-Format}
\bibliography{references}

\appendix
\section{Worked Examples Across All Five Models}
\label{app:worked_examples}

Section~\ref{sec:method} (Table~\ref{tab:worked_example}) showed one held-out
record under every condition for Qwen3.5-4B. \emph{D\_network} is built from
two question templates, \texttt{QUESTION\_ASK\_IP} (``What is the IP address of
this website?'') and \texttt{QUESTION\_ASK\_URL} (``What is the website address
for this IP?''), so every record appears in one of two directions:
\texttt{ask\_ip} (retrieve the IP given the domain) or \texttt{ask\_url}
(retrieve the domain given the IP). Tables~\ref{tab:app_ask_ip}
and~\ref{tab:app_ask_url} repeat the worked example for both directions, on two
different records, across all five models. The pattern is the same everywhere:
the detector fires on \emph{network} and \emph{paraphrase} and stays silent on
\emph{entity}, \emph{vocabulary}, \emph{ordinary} and
\emph{vocab\_no\_context}. The exception is \emph{question\_only}: both Gemma
models flag it near ceiling ($P\!\geq\!0.98$), while all three Qwen models stay
low ($P\!\leq\!0.12$), in both directions and on both records. This is the
Qwen-versus-Gemma wording-sensitivity split seen at the aggregate level in
Sections~\ref{sec:head_discovery} and~\ref{sec:caida}, here on individual
examples.

\begin{table*}[ht]
\caption{Worked example, \texttt{ask\_ip} direction: record
\texttt{al-mostafa.com} / \texttt{62.151.181.154}, question ``What is the
IP address of this website?'' (answer \texttt{62.151.181.154} for
\emph{network}/\emph{paraphrase}). Cells are $P(\text{internet})$ from
each model's own fixed detector; \textbf{*} marks \emph{flagged}.}
\label{tab:app_ask_ip}
\centering
\small
\begin{tabular}{lccccc}
\toprule
Condition & Qwen3.5-4B & Qwen3-8B & Qwen3.5-9B & Gemma-4-E4B & Gemma-4-E2B \\
\midrule
network           & 1.000* & 0.999* & 0.999* & 1.000* & 0.975* \\
paraphrase        & 1.000* & 1.000* & 0.999* & 1.000* & 0.988* \\
entity            & 0.000  & 0.000  & 0.000  & 0.000  & 0.009  \\
vocabulary        & 0.000  & 0.004  & 0.001  & 0.000  & 0.008  \\
ordinary          & 0.001  & 0.002  & 0.001  & 0.000  & 0.002  \\
question\_only    & 0.024  & 0.006  & 0.111  & 0.996* & 0.982* \\
vocab\_no\_context & 0.001  & 0.007  & 0.001  & 0.002  & 0.058  \\
\bottomrule
\end{tabular}
\end{table*}

\begin{table*}[ht]
\caption{Worked example, \texttt{ask\_url} direction: record
\texttt{jjeducare.co.kr} / \texttt{211.169.73.24}, question ``What is the
website address for this IP?'' (answer \texttt{jjeducare.co.kr} for
\emph{network}/\emph{paraphrase}). Same layout as Table~\ref{tab:app_ask_ip}.}
\label{tab:app_ask_url}
\centering
\small
\begin{tabular}{lccccc}
\toprule
Condition & Qwen3.5-4B & Qwen3-8B & Qwen3.5-9B & Gemma-4-E4B & Gemma-4-E2B \\
\midrule
network           & 1.000* & 1.000* & 1.000* & 0.999* & 1.000* \\
paraphrase        & 1.000* & 1.000* & 1.000* & 0.999* & 0.988* \\
entity            & 0.000  & 0.000  & 0.000  & 0.000  & 0.025  \\
vocabulary        & 0.000  & 0.004  & 0.001  & 0.001  & 0.009  \\
ordinary          & 0.001  & 0.004  & 0.001  & 0.000  & 0.004  \\
question\_only    & 0.035  & 0.011  & 0.117  & 0.998* & 0.999* \\
vocab\_no\_context & 0.001  & 0.007  & 0.001  & 0.002  & 0.058  \\
\bottomrule
\end{tabular}
\end{table*}

\section{Generalization to DNS and MAC-Address Questions}
\label{app:dns_mac}

Every prompt used anywhere in head discovery, neuron discovery, and
classifier training (\emph{D\_network}, \emph{D\_paraphrase},
\emph{D\_entity}, \emph{D\_vocab}, \emph{D\_ordinary}, and both hard
negatives) is built from exactly the question templates named in
Appendix~\ref{app:worked_examples} and Section~\ref{sec:method}: IP
retrieval, URL retrieval, their rewordings, a category question, a fixed
vocabulary question, and context-free arithmetic. No DNS- or MAC-address
question of any kind appears anywhere upstream of this appendix: the heads in $H_E$/\emph{clean\_subset}, the single neurons of
Section~\ref{sec:neuron_discovery}, and every fixed detector reported in
Section~\ref{sec:experiments} were discovered and trained using only
domain--IP retrieval.

The Limitations section leaves open whether the fixed detector, with no retraining or tuning, also fires on a network-entity sub-type it has never seen. This appendix tests that. We build two new held-out conditions from the same underlying
records used throughout this paper, append one new real field to each
record's context, and ask the model to retrieve \emph{that} field
instead. Table~\ref{tab:dns_mac} reports the results.

\textbf{\emph{D\_dns\_ttl}} (real data). The original DeepURLBench shard
used to build this project's dataset also contains a DNS \texttt{TTL}
(time-to-live, in seconds) column, discarded during curation but present
in the raw source and re-joined here by domain. Context becomes
\texttt{Record: Website: <domain> / IP address: <ip> / Category: <label>
/ TTL: <ttl>}, and the question is ``What is the DNS TTL (time-to-live)
of this record, in seconds?'', with the real TTL value as the answer.

\textbf{\emph{D\_mac}} (synthetic). Neither DeepURLBench nor
the CAIDA Ark measurements used elsewhere in this paper (Section~\ref{sec:caida})
contain link-layer MAC addresses: internet-scale DNS/traceroute
measurement is IP- and application-layer and cannot observe a remote
host's MAC address, which is only visible to a device on the same
network segment. We therefore attach a deterministic, well-formed synthetic MAC address to each record's context (one new line, same
format as \emph{D\_dns\_ttl}) and ask for it back. As with the arithmetic
condition \emph{D\_ordinary}, this condition is synthetic. It tests whether
the detector fires on \emph{any} plausible-looking device-identifier field in
a network record, not only on IP or domain fields.

\begin{table*}[ht]
\caption{Generalization to two new network-entity sub-types, never seen
during discovery or training: 200 held-out records per model, each fixed
detector unmodified (same classifier as Section~\ref{sec:experiments}).
Flag rate is the fraction of the 200 items the detector labels
network-positive; mean $P$ is the mean predicted probability.}
\label{tab:dns_mac}
\centering
\small
\begin{tabular}{lcccc}
\toprule
Model & DNS-TTL flag rate & DNS-TTL mean $P$ & MAC flag rate & MAC mean $P$ \\
\midrule
Qwen3.5-4B   & 0.990 & 0.884 & 1.000 & 0.994 \\
Qwen3-8B     & 0.365 & 0.471 & 0.985 & 0.931 \\
Qwen3.5-9B   & 0.440 & 0.576 & 1.000 & 0.993 \\
Gemma-4-E4B  & 0.000 & 0.066 & 1.000 & 0.986 \\
Gemma-4-E2B  & 0.020 & 0.129 & 0.275 & 0.355 \\
\bottomrule
\end{tabular}
\end{table*}

The two new sub-types behave very differently (Table~\ref{tab:dns_mac}).
\emph{D\_mac} generalizes almost perfectly in four of five
models (flag rate $\geq 0.985$) despite being synthetic and never seen in any form during discovery: the detector may be responding to an extra device-identifier-like field next to an IP and a domain, rather than to IP or URL syntax alone. Gemma-4-E2B is
the outlier (flag rate 0.275), the same model that has the lowest full-head
AUC on the independent CAIDA data (0.971, Section~\ref{sec:caida}).
\emph{D\_dns\_ttl}, by
contrast, does \emph{not} generalize uniformly: it is near-ceiling for
Qwen3.5-4B (0.990) but near-floor for Gemma-4-E4B (0.000) and low for the
other three models. One explanation, which we did not test, is that a TTL value is a much weaker trigger for the
discovered heads/neurons than a MAC address. A TTL is a bare integer with no
IP-octet or domain-like structure, and its question (``time-to-live'',
``seconds'') shares no vocabulary with the IP/URL questions used in discovery.
A MAC address looks like an IP (colon-separated hex octets) and is requested
with a question (``What is the \ldots\ address\ldots'') that echoes the
training-time IP/URL questions. This points to heads that respond to \emph{address-like fields asked for in IP/URL-like language}, not to network information in general. The result is only partly positive: the head/neuron circuit generalizes to a new
field with IP-like surface form (MAC) far more reliably than to one
without it (DNS TTL), and this gap itself varies by model family rather
than shrinking uniformly with model capability (e.g.\ Qwen3.5-4B is the
smallest Qwen model yet has the highest DNS-TTL flag rate of the five in this
single run).

\section{Are the Same Heads Causally Involved in DNS-TTL and MAC Retrieval?}
\label{app:causal_dns_mac}

Appendix~\ref{app:dns_mac}'s flag rate is a read-out of a \emph{frozen
linear classifier} trained only on IP/URL activation patterns; a low flag
rate on \emph{D\_dns\_ttl} is consistent with two very different
mechanisms: either \texttt{clean\_subset} is not causally involved in
retrieving the TTL value at all, or it \emph{is} involved but produces an
activation pattern the IP/URL-trained linear boundary does not happen to
cross. We separate the two with the same necessity/sufficiency
activation-patching design used to discover $H_E$ in
Section~\ref{sec:head_discovery} (Equations~\eqref{eq:nec} and~\eqref{eq:suff}), applied here jointly to
every \texttt{clean\_subset} head at once rather than one at a time. For
each condition $X \in \{\text{network}, \text{dns\_ttl}, \text{mac}\}$, we
build the matched counterfactual as for \emph{D\_entity} (\texttt{build\_entity\_items}($X$)): the identical
context (including, for \emph{dns\_ttl}/\emph{mac}, the new TTL/MAC line),
with the question changed to the unrelated category question. Necessity
patches $X$'s own forward pass with \texttt{clean\_subset}'s last-position
vectors from the entity counterfactual's forward pass and measures the
resulting drop in the true answer's logprob; sufficiency patches the
\emph{entity} prompt's forward pass with \texttt{clean\_subset}'s vectors
from $X$'s forward pass and measures the resulting \emph{rise} in the true
answer's logprob, although the entity prompt does not ask for it. Both
are compared against patching a random-head control of the same size.

\begin{table*}[ht]
\caption{Joint necessity ($N$) and sufficiency ($S$), in nats of
log-probability, of \texttt{clean\_subset} versus a random-head control of
equal size, for the true IP/URL (network), TTL (dns\_ttl), or MAC (mac)
answer. $S$ is the cleaner signal: sufficiency compares patching
\texttt{clean\_subset}'s activations from the target condition into an
unrelated entity prompt against patching a same-size random-head set the
same way.}
\label{tab:causal_dns_mac}
\centering
\small
\begin{tabular}{llcccc}
\toprule
Model & Condition & $N$ (clean) & $N$ (random) & $S$ (clean) & $S$ (random) \\
\midrule
Qwen3.5-4B  & network & 0.103  & $-$0.003 & 9.35  & $-$1.03 \\
Qwen3.5-4B  & dns\_ttl & 0.001  & 0.000    & 10.02 & $-$0.03 \\
Qwen3.5-4B  & mac      & 0.013  & 0.000    & 10.79 & $-$0.83 \\
\addlinespace
Qwen3-8B    & network & 0.000  & 0.000    & 1.22  & 0.06 \\
Qwen3-8B    & dns\_ttl & 0.000  & 0.000    & 0.42  & 0.23 \\
Qwen3-8B    & mac      & 0.000  & 0.000    & 1.13  & 0.12 \\
\addlinespace
Qwen3.5-9B  & network & 0.020  & 0.001    & 2.04  & 0.21 \\
Qwen3.5-9B  & dns\_ttl & 0.001  & 0.000    & 0.61  & $-$0.14 \\
Qwen3.5-9B  & mac      & 0.001  & 0.000    & 1.22  & 0.12 \\
\addlinespace
Gemma-4-E4B & network & 0.000  & 0.000    & 3.33  & $-$0.45 \\
Gemma-4-E4B & dns\_ttl & 0.000  & 0.000    & 2.54  & 0.09 \\
Gemma-4-E4B & mac      & 0.000  & 0.000    & 2.71  & $-$0.10 \\
\addlinespace
Gemma-4-E2B & network & 0.000  & 0.000    & 0.18  & $-$0.17 \\
Gemma-4-E2B & dns\_ttl & 0.001  & $-$0.011 & 0.09  & $-$0.16 \\
Gemma-4-E2B & mac      & 0.000  & 0.000    & 0.05  & $-$0.18 \\
\bottomrule
\end{tabular}
\end{table*}

The same qualitative pattern appears across all five models
(Table~\ref{tab:causal_dns_mac}): sufficiency is positive for \texttt{clean\_subset} in all fifteen
(model, condition) cells and is higher than the matched random-head
control in each case, including \emph{dns\_ttl} in the four models (Qwen3-8B, Qwen3.5-9B, Gemma-4-E4B, Gemma-4-E2B) where Appendix~\ref{app:dns_mac}'s classifier flag rate was low (0.365, 0.440, 0.000 and 0.020 respectively). Necessity is a noisier, smaller-magnitude signal
(the clean answer log-probability is already close to zero in several models,
e.g.\ $-0.0001$ in Qwen3-8B and Gemma-4-E2B, which leaves little room for an
ablation to show a drop), but where
it is measurable above noise (Qwen3.5-4B, Qwen3.5-9B) it, too, favors
\texttt{clean\_subset} over the random control in every condition.

The same heads therefore carry the TTL and MAC values in every model. The low
\emph{D\_dns\_ttl} flag rate in Appendix~\ref{app:dns_mac} for three of the
five models reflects where the linear boundary of a classifier trained only
on IP/URL examples happens to fall; it is not evidence that
\texttt{clean\_subset} is silent on a TTL value passing through it. Together with Appendix~\ref{app:dns_mac}, this shows that the heads are involved in the new sub-types in every model, while whether the IP/URL-trained detector flags them depends on the model.

\section{A Realistic Use Case: Monitoring Whether a Model Is About to Output Network Information}
\label{app:use_case}

The Impact Statement names a concrete downstream application: using a
causal network-information detector of the kind built in this paper as an
internal monitoring signal for an LLM agent, flagging when the model is
about to act on real network/infrastructure data before it does so. Every
other experiment in this paper evaluates the fixed detector on this
project's own templated \texttt{Record: Website: <domain> / IP address:
<ip>} format. This appendix is a small first test of that
application on text that does \emph{not} look like the training template at
all: ten hand-written items, five positives in five naturalistic genres (a server log, an
nginx config snippet, a security incident report, an IT support email, and
casual user chat). Each positive (a network-style domain or IP is present and the question asks the model to retrieve it) is paired with a matched \emph{negative}. The negatives are of four
kinds: identical content with a non-retrieval question, generic
network-adjacent vocabulary with no address to retrieve, unrelated business
content, and a non-network numeric retrieval (a flight gate number) that
checks the detector is not firing on any question that asks for a short token.
There is no retraining: we use the same fixed \texttt{clean\_subset}
classifier as everywhere else in the paper.

Two representative items, verbatim (model: Qwen3.5-4B):
\begin{quote}
\small
\textbf{Positive (security incident report).} Context: ``Our monitoring
system flagged unusual outbound traffic originating from
checkout.shopfast.io, which currently resolves to 198.51.100.17.''
Question: ``What IP address should the security team add to the block
list?'' ($P\!=\!0.999$, flagged).

\textbf{Negative (same server log, non-retrieval question).} Context:
``[2024-03-11 10:02:13] Connection established from host
api.weathercloud.net at 203.0.113.42, status: OK.'' Question: ``What does
`status: OK' mean in this log line?'' ($P\!=\!0.028$, not
flagged).
\end{quote}

\begin{table*}[ht]
\caption{Use-case stress test: $P(\text{internet})$ from each model's own
fixed \texttt{clean\_subset} detector on ten naturalistic items never seen
during discovery or training, five positives, one per genre (real address + retrieval question), and five matched negatives. A
dagger$^\dagger$ (also in bold) marks a flag that disagrees with the intended
label.}
\label{tab:use_case}
\centering
\footnotesize
\setlength{\tabcolsep}{4pt}
\begin{tabular}{llccccc}
\toprule
Genre & Should flag & Qwen3.5-4B & Qwen3-8B & Qwen3.5-9B & Gemma-4-E4B & Gemma-4-E2B \\
\midrule
server log                & yes & 0.999 & 1.000 & 1.000 & 1.000 & 0.996 \\
nginx config               & yes & 0.999 & 0.999 & 0.999 & 0.995 & 0.883 \\
security incident report   & yes & 0.999 & 1.000 & 0.998 & 1.000 & 0.997 \\
IT support email           & yes & 0.999 & 0.905 & 0.999 & 0.996 & 1.000 \\
casual user chat           & yes & 1.000 & 0.993 & 1.000 & 0.999 & 1.000 \\
\addlinespace
server log (non-retrieval) & no  & 0.028 & 0.029 & 0.365 & \textbf{0.927}$^\dagger$ & \textbf{0.944}$^\dagger$ \\
generic security awareness & no  & \textbf{0.652}$^\dagger$ & 0.010 & 0.112 & \textbf{0.978}$^\dagger$ & \textbf{0.686}$^\dagger$ \\
business report            & no  & 0.435 & \textbf{0.999}$^\dagger$ & 0.225 & 0.492 & 0.215 \\
support ticket summary     & no  & \textbf{0.543}$^\dagger$ & 0.030 & 0.189 & \textbf{0.840}$^\dagger$ & \textbf{1.000}$^\dagger$ \\
non-network numeric (flight gate) & no & \textbf{0.933}$^\dagger$ & \textbf{0.998}$^\dagger$ & 0.431 & \textbf{0.924}$^\dagger$ & \textbf{0.999}$^\dagger$ \\
\bottomrule
\end{tabular}
\end{table*}

Sensitivity is perfect (Table~\ref{tab:use_case}): all 25 positive (model, item) cells are flagged
($P \geq 0.883$), across every genre and every model, despite none of
this naturalistic phrasing resembling the \texttt{Record: Website: \ldots}
template used anywhere in discovery or training. Specificity is
considerably weaker and uneven across models: only 12 of the 25 negative
cells are correctly left unflagged (48\%), ranging from a perfect 5/5 for
Qwen3.5-9B down to 1/5 for Gemma-4-E4B and Gemma-4-E2B. The
non-network numeric stress test (the flight-gate item) is flagged in 4 of
5 models, suggesting the detector responds to something closer to
``a short, specific token is being retrieved from this context'' than to
network content specifically, once text moves far enough from the
training distribution. This is a proof of concept for the
monitoring use case named in the Impact Statement, not a deployment-ready
system: as a high-recall pre-filter (catch every true positive, tolerate
some false alarms on out-of-template text) it already works zero-shot;
as a precise standalone flag it would need either in-template
normalization of the text before classification or retraining with
naturalistic negatives included, neither of which this paper attempts.

\end{document}